\documentclass[letterpaper, 10 pt, conference]{ieeeconf}  

\IEEEoverridecommandlockouts                              

\usepackage{float}
\usepackage{graphicx}
\usepackage{graphics} 
\usepackage{amsmath} 
\usepackage{amssymb}  
\usepackage{cite}
\usepackage[hidelinks]{hyperref}
\usepackage{booktabs}
\usepackage{balance}
\usepackage{multirow}
\usepackage{adjustbox}
\usepackage{subcaption}

\title{\LARGE \bf
Quaternion Approximate Networks for Enhanced Image Classification and Oriented Object Detection}

\author{Bryce Grant$^{1}$ and Peng Wang$^{2}$%
\thanks{The author from $^{1}$ is with the Department of Electrical, Systems and Computer Engineering; Case Western Reserve University
$^{2}$ is with Department of Mechanical and Aerospace Engineering, Case Western Reserve University, Cleveland, Ohio, USA}%
\thanks{E-Mail: {\tt\small bag100@case.edu and pxw206@case.edu}}
\thanks{Code available at: \url{https://cwru-aism.github.io/QUANpaper/}}
\thanks{This work is supported by the National Science Foundation under grant No. 2024614.}
}

\begin{document}

\maketitle
\thispagestyle{empty}
\pagestyle{empty}

\begin{abstract}
This paper introduces Quaternion Approximate Networks (QUAN), a novel deep learning framework that leverages quaternion algebra for rotation equivariant image classification and object detection. Unlike conventional quaternion neural networks attempting to operate entirely in the quaternion domain, QUAN approximates quaternion convolution through Hamilton product decomposition using real-valued operations. This approach preserves geometric properties while enabling efficient implementation with custom CUDA kernels. We introduce Independent Quaternion Batch Normalization (IQBN) for training stability and extend quaternion operations to spatial attention mechanisms. QUAN is evaluated on image classification (CIFAR-10/100, ImageNet), object detection (COCO, DOTA), and robotic perception tasks. In classification tasks, QUAN achieves higher accuracy with fewer parameters and faster convergence compared to existing convolution and quaternion-based models. For objection detection, QUAN demonstrates improved parameter efficiency and rotation handling over standard Convolutional Neural Networks (CNNs) while establishing the SOTA for quaternion CNNs in this downstream task. These results highlight its potential for deployment in resource-constrained robotic systems requiring rotation-aware perception and application in other domains. 
\end{abstract}



\begin{figure}[h]
    \centering
    \begin{subfigure}[b]{.5\columnwidth}
        \includegraphics[width=\columnwidth]{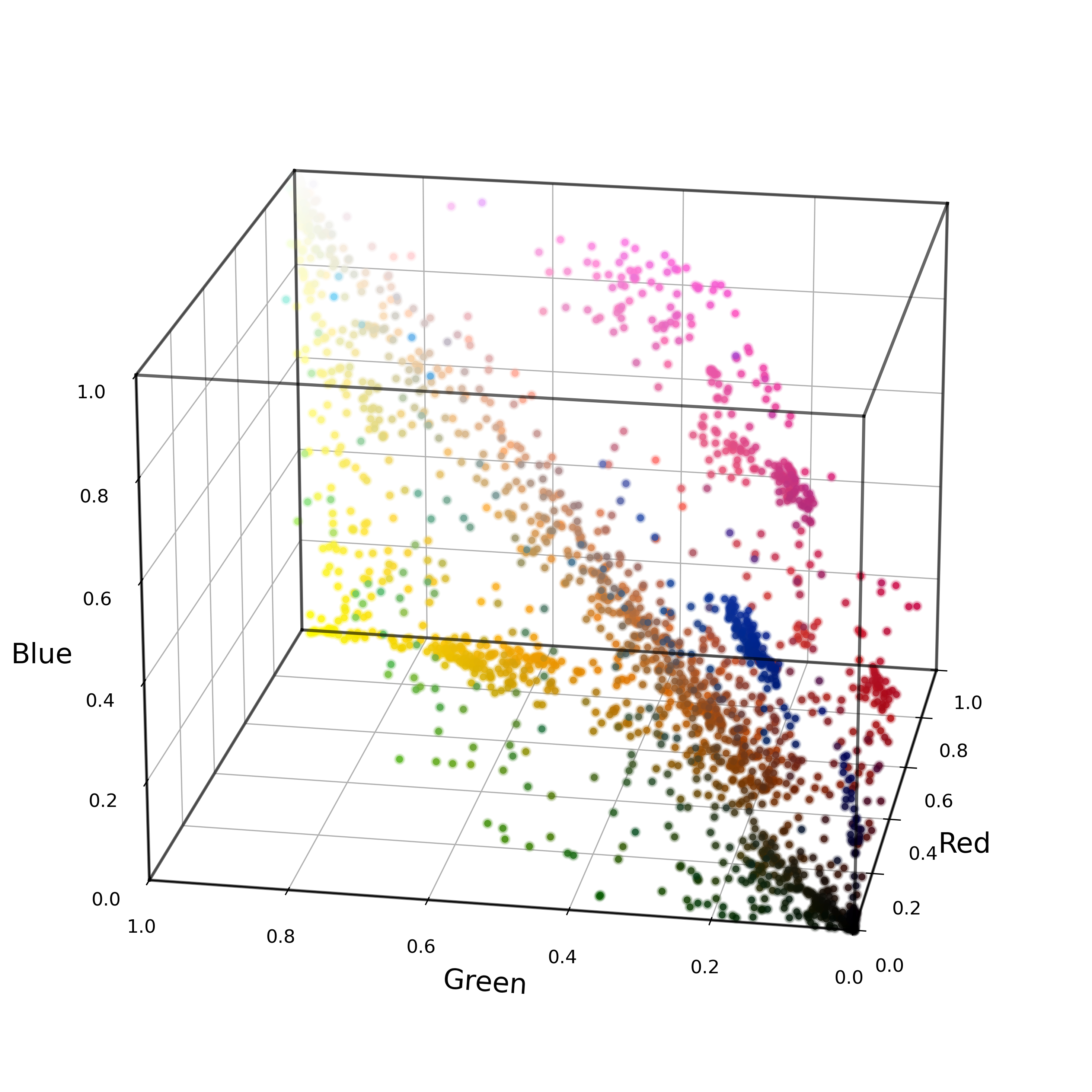}
        \caption{RGB color space}
        \label{fig:a}
    \end{subfigure}
    \hfill
    \begin{subfigure}[b]{.5\columnwidth}
        \includegraphics[width=\columnwidth]{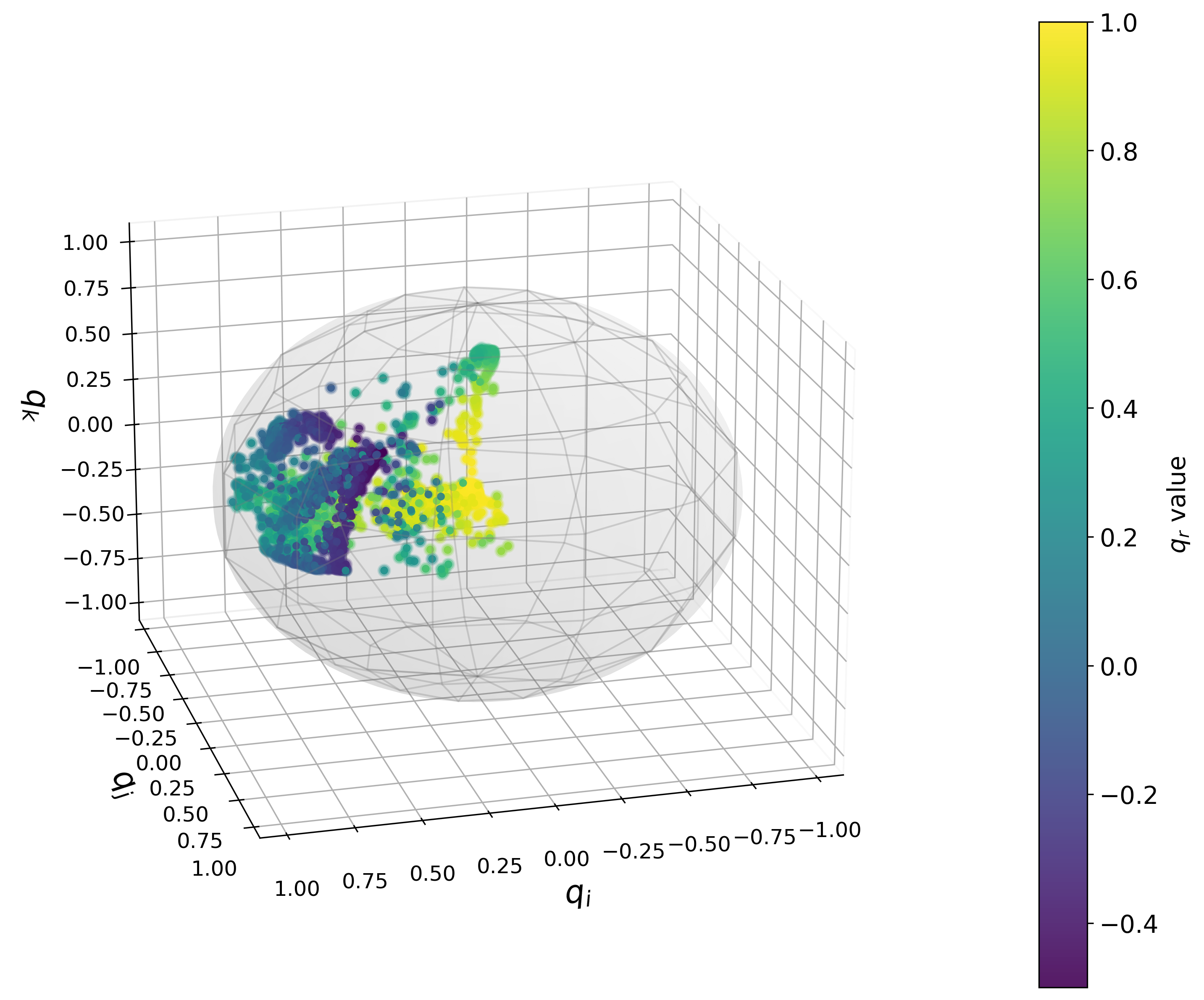}
        \caption{Quaternion Space}
        \label{fig:b}
    \end{subfigure}
    \caption{Poincaré mapping from RGB color space to quaternion space.}
    \label{fig:poincare}
\end{figure}

\section{INTRODUCTION}
Vision-based robotic assembly and manipulation tasks represent a critical challenge in modern robotics, which requires precise object detection, pose estimation, and spatial reasoning capabilities. Conventional computer vision approaches typically struggle with object orientation, a fundamental requirement for robotic interaction with real-world objects. Although deep learning has revolutionized object detection through frameworks such as YOLO \cite{yolov11}, Faster R-CNN \cite{fastcnn}, and DETR \cite{endtoend}, most approaches rely on axis-aligned bounding boxes that discard crucial orientation information. This limitation significantly affects performance in robotic applications where grasp planning and assembly operations are highly dependent on accurate object orientation \cite{densefusion}. 

Recent advances in vision-language models and self-supervised learning have transformed the computer vision landscape. DINO \cite{dino} and other contrastive learning approaches \cite{simclr} have demonstrated improved performance by learning from unlabeled data, while Mixture-of-Experts (MoE) architectures \cite{switch} achieve state-of-the-art results through adaptive computation pathways. However, these sophisticated models come with substantial computational requirements, often containing hundreds of millions of parameters. Despite their impressive general-purpose capabilities, they lack the inherent mathematical structures to efficiently represent and process rotational information, a critical requirement for robotic manipulation tasks.

Meanwhile, specialized approaches for oriented object detection \cite{r3det, Xu21} have emerged to address rotation-aware perception needs. These methods typically use complex architectures and loss functions to estimate object orientation, but they often require substantial computational resources. This computational burden presents a significant barrier for deployment in resource-constrained robotic systems that require real-time performance.

Quaternion algebra offers  a powerful mathematical framework for encoding and manipulating three-dimensional rotations \cite{hamilton1844,kuipers1999quaternions}. Originally developed by Hamilton in 1843, quaternions avoid the singularities associated with Euler angles (gimbal lock) and provide a more compact representation compared to rotation matrices. This mathematical elegance has inspired researchers to explore quaternion-based deep learning approaches. Gaudet and Maida \cite{DQN} established foundational concepts in deep quaternion networks, introducing quaternion backpropagation and initialization schemes for image classification, while Parcollet et al. \cite{QRNN} demonstrated the parameter efficiency of quaternion neural networks for audio processing tasks. Zhu et al.~\cite{QCNN} further extended this to image denoising tasks, showing quaternion CNNs could achieve comparable performance to standard CNNs with significantly fewer parameters.

Despite these advances, existing quaternion neural networks face substantial implementation challenges. Pure quaternion networks require specialized mathematical operations that are poorly supported in mainstream deep learning frameworks, leading to significant computational overhead. Gomez and Gershenson \cite{gomez23} highlighted fundamental limitations such as the non-existence of bounded non-constant analytic quaternion functions, constraining the design of activation functions and complicating gradient computations. These practical limitations have prevented quaternion networks from being deployed in real-world robotic applications despite their theoretical advantages.

This paper introduces Quaternion Approximate Networks (QUAN), a novel approach that preserves the rotation equivariance benefits of quaternion representations while maintaining computational efficiency. Rather than performing operations in the quaternion domain directly, QUAN implements quaternion convolution through Hamilton product approximation using standard real-valued operations. This approach allows leveraging optimized tensor operations in existing deep learning frameworks while retaining the representational advantages of quaternion.

The key contributions are as follows:
\begin{itemize}
\item A quaternion approximation approach that implements Hamilton-based mixing using real-valued operations, maintaining rotation equivariance while significantly reducing computational complexity compared to previous quaternion neural networks.
\item Novel quaternion-aware architectural components, including Independent Quaternion Batch Normalization (IQBN) and Quaternion Partial Spatial Attention (QC2PSA), that enhance model performance while preserving quaternion structure.
\item State-of-the-art performance for quaternion networks on image classification tasks (CIFAR-10, CIFAR-100, ImageNet) with significantly reduced parameter counts compared to both standard CNNs and existing quaternion approaches.
\item A systematic quaternion adaptation of YOLO architecture components (QC3k2, QSPPF, QC2PSA) for oriented bounding box detection, benchmarked on a custom robotic manipulation dataset.
\item A novel orientation-aware loss function unifying quaternion angular metrics with traditional detection objectives, enabling end-to-end learning of both object location and orientation in quaternion space.
\end{itemize}

QUAN represents the first quaternion-inspired architecture that successfully bridges the gap between theoretical quaternion advantages and practical deployment needs for robotic perception systems. By combining parameter efficiency with rotation equivariance properties, this approach enables robust object detection and pose estimation for robotic manipulation tasks while maintaining suitable for real-time applications.

\section{Related Works}
\subsection{General Object Detection and Pose Estimation}
Object detection forms the foundation of many robotic vision systems. Landmark architectures like YOLO \cite{yolov11}, Faster R-CNN \cite{fastcnn}, and DETR \cite{endtoend} have established new performance benchmarks but primarily focus on axis-aligned bounding boxes, which limit their utility for robotic manipulation tasks where orientation is crucial. For robotic applications, Wang et al. [4] developed DenseFusion, by fusing RGB and depth information, while Hodan et al. \cite{BOP} introduced a benchmark for 6D object pose estimation in cluttered scenes while Kleeberger et al. \cite{bin} introduced a benchmark for industrial bin picking, highlighting the challenges of accurate orientation estimation in realistic settings. These approaches, while effective, often separate the detection and pose estimation steps rather than handling them in a unified framework.

Specialized methods for oriented object detection have emerged to address these limitations. Yang et al. \cite{r3det} proposed R3Det, which refines feature maps for detecting rotated objects, while Xu et al. \cite{Xu21} introduced the gliding vertex mechanism to generate arbitrarily oriented bounding boxes. However, these approaches typically require complex architectural modifications and substantially increased computational resources compared to standard detectors.

\subsection{Quaternion-Based Methods}
Quaternions provide a compact and singularity-free representation for 3D rotations \cite{hamilton1844,kuipers1999quaternions}. Gaudet and Maida \cite{DQN} pioneered quaternion neural networks by establishing fundamental operations, including weight initialization and backpropagation. Parcollet et al. \cite{QRNN} extended this work to recurrent networks and demonstrated parameter efficiency in speech processing tasks. Zhu et al. \cite{QCNN} applied quaternion convolutions to image processing, achieving comparable performance to standard CNNs with fewer parameters. Comminiello et al. \cite{qsound} represented microphone in their spherical harmonics form, which enabled the use of quaternion networks for the detection of 3D sound events. Grassucci et al. \cite{Grassucci_2021} investigated quaternion generative adversarial networks, to obtain better FID scores than real-valued GANs. These works showed that quaternion CNNs could achieve comparable or even superior performance to standard CNNs with fewer parameters.

However, quaternion neural networks face significant implementation challenges. Gomez and Gershenson \cite{gomez23} identified limitations including higher computational complexity and the non-existence of bounded non-constant analytic quaternion functions, which constrains activation function design. These practical challenges have prevented widespread adoption despite theoretical advantages.

\subsection{Quaternion Pose Estimation}
Rotation equivariance is particularly valuable for robotic perception. Hsu et al. ~\cite{quatnet2019} introduced QuatNet, using quaternions for continuous head pose estimation with a multi-regression loss function. He et al.~\cite{pvn3d2020} introduced PVN3D, a deep point-wise 3D keypoints voting network for 6DoF pose estimation using quaternions for rotation representation. Despite these advances, no previous work has successfully applied quaternion-based approaches to oriented object detection in practical robotic applications, which is the focus of this paper.

\subsection{Alternative Rotation Equivariant Architectures}

While our work focuses on quaternion-based approaches, several other frameworks achieve rotation equivariance through different mechanisms. Cohen and Welling~\cite{cohen2016group} introduced Group Equivariant CNNs (G-CNNs) that encode symmetries using group theory, ensuring equivariance to discrete rotations. Weiler and Cesa~\cite{weiler2019general} extended this with steerable CNNs using continuous groups, achieving equivariance to arbitrary rotations. Marcos et al.~\cite{marcos2017rotation} proposed Rotation Equivariant Vector Field Networks that explicitly encode angles in the network architecture.

These approaches, require fundamental architectural changes and specialized implementations. Harmonic Networks~\cite{worrall2017harmonic} use circular harmonics to achieve rotation equivariance, but suffer from increased computational complexity. Oriented Response Networks~\cite{zhou2017oriented} actively rotate filters during convolution, resulting in significant overhead.

In contrast, QUAN leverages quaternion algebra's inherent rotation properties within standard deep learning frameworks, achieving rotation awareness without the architectural complexity or computational burden of pure group-theory approaches.

\section{Methodology}
\begin{figure*}[h!]
    \centering
    \includegraphics[width=\textwidth]{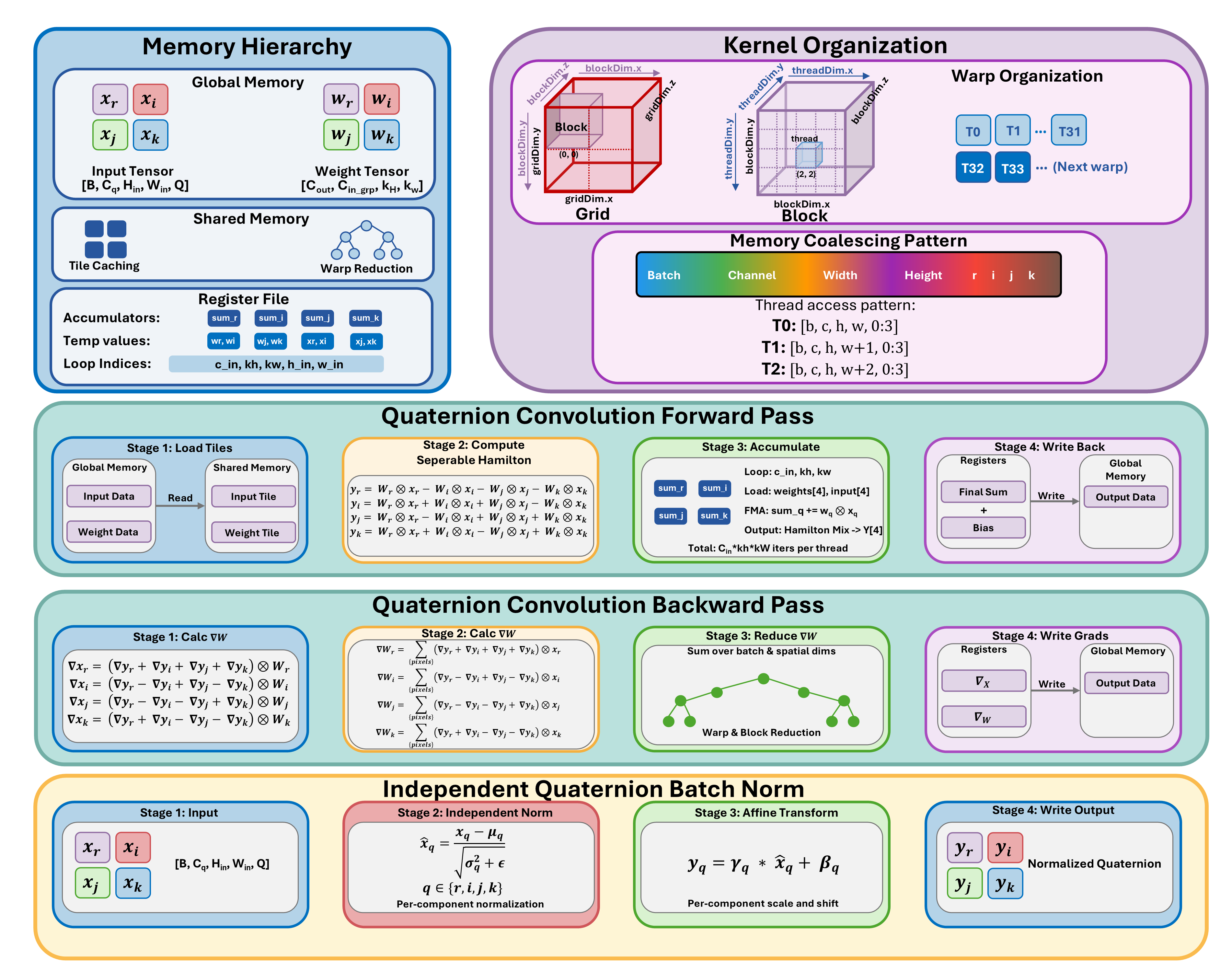}
    \caption{CUDA kernel implementation for quaternion convolution}
    \label{fig:cuda_kernel}
\end{figure*}

\subsection{Quaternion Representation and Mapping} \label{subsection}

The quaternion number system extends complex numbers to four dimensions, represented as:

\begin{equation}
q = q_r + q_i\mathbf{i} + q_j\mathbf{j} + q_k\mathbf{k}
\end{equation}

where $q_r, q_i, q_j, q_k \in \mathbb{R}$, and $\mathbf{i}, \mathbf{j}, \mathbf{k}$ are the fundamental quaternion units satisfying:

\begin{equation}
\mathbf{i}^2 = \mathbf{j}^2 = \mathbf{k}^2 = \mathbf{i}\mathbf{j}\mathbf{k} = -1
\end{equation}

The Hamilton product between two quaternions $p = p_r + p_i\mathbf{i} + p_j\mathbf{j} + p_k\mathbf{k}$ and $q = q_r + q_i\mathbf{i} + q_j\mathbf{j} + q_k\mathbf{k}$ is defined as:
\begin{equation}
\begin{aligned}
p \otimes q &= (p_r q_r - p_i q_i - p_j q_j - p_k q_k) \\
&+ (p_r q_i + p_i q_r + p_j q_k - p_k q_j)\mathbf{i} \\
&+ (p_r q_j - p_i q_k + p_j q_r + p_k q_i)\mathbf{j} \\
&+ (p_r q_k + p_i q_j - p_j q_i + p_k q_r)\mathbf{k}
\end{aligned}
\end{equation}

For image processing applications, RGB data must be mapped into quaternion space. Several mapping strategies have been explored, as explained in the Appendix \autoref{app:mapping_strategies}, with evaluation demonstrating that the Poincaré mapping \cite{hyperbolic, poincare}, which embeds the RGB cube within the unit quaternion sphere shown in Fig. \ref{fig:poincare}, provided superior performance:

\begin{equation}
\begin{split}
\mathbf{norm} &= \sqrt{r^2 + g^2 + b^2} \\
q_r &= \frac{1-\mathbf{norm^2}}{1+ \mathbf{norm^2}} \\
q_i &= \frac{2r}{1+ \mathbf{norm^2}} \\
q_j &= \frac{2g}{1+ \mathbf{norm^2}} \\
q_k &= \frac{2b}{1+ \mathbf{norm^2}} \\
q &= [q_r, q_i, q_j, q_k]
\end{split}
\end{equation}


Poincaré mapping embeds RGB values into the into the Poincaré ball model of hyperbolic space, where the distance between points increases exponentially as they approach the boundary of the unit ball. This property allows the quaternion representation to better preserve hierarchical relationships in the color space compared to Euclidian embeddings. 

Our approach is an approximation for two reasons: (1) we decompose the full quaternion convolution into separable operations rather than computing all 16 terms of the Hamilton product, reducing computational complexity from $\mathcal{O}(16CHW)$ to $\mathcal{O}(4CHW)$, and (2) we relax the unit quaternion constraint $\|q\|=1$ during training, instead using Independent Quaternion Batch Normalization (IQBN) to maintain numerical stability. We found that enforcing strict constraints through Riemannian optimization \cite{kochurov2020geooptriemannianoptimizationpytorch} degraded performance, suggesting that our formulation better captures features while maintaining the quaternion structure.

\subsection{Hamilton-Based Quaternion Approximation}

Traditional quaternion convolution applies the Hamilton product between quaternion-valued kernels and features. For quaternion convolution with kernel ${W}$ and input feature map ${X}$, where each quaternion has components $(r, i, j, k)$, the full Hamilton product expansion is:
\begin{equation}
\begin{aligned} \label{eq:5}
{Y} &= ({W}_r \otimes {X}_r - {W}_i \otimes {X}_i - {W}_j \otimes {X}_j - {W}_k \otimes {X}_k) \\
&+ ({W}_r \otimes {X}_i + {W}_i \otimes {X}_r + {W}_j \otimes {X}_k - {W}_k \otimes {X}_j)\mathbf{i} \\
&+ ({W}_r \otimes {X}_j - {W}_i \otimes {X}_k + {W}_j \otimes {X}_r + {W}_k \otimes {X}_i)\mathbf{j} \\
&+ ({W}_r \otimes {X}_k + {W}_i \otimes {X}_j - {W}_j \otimes {X}_i + {W}_k \otimes {X}_r)\mathbf{k}
\end{aligned}
\end{equation}
where $\otimes$ denotes standard convolution. The standard Hamilton matrix can be derived as:
\begin{equation} \label{eq:6}
\mathbf{W} \otimes \mathbf{X} = 
\begin{bmatrix} 
w_r & -w_i & -w_j & -w_k \\
w_i & w_r & -w_k & w_j \\
w_j & w_k & w_r & -w_i \\
w_k & -w_j & w_i & w_r
\end{bmatrix}
\begin{bmatrix} x_r \\ x_i \\ x_j \\ x_k \end{bmatrix}
\end{equation}

To approximate the quaternion algebra, QUAN decomposes the left-hand operation into four separate convolutions, a method inspired by \cite{Zhou}. For an input tensor $X \in \mathbb{R}^{B \times C \times H \times W \times Q}$ and convolutional kernel $W$, the separable QUAN convolution forward pass is defined as:

\begin{equation}
\begin{split}
y_r &= W_r \otimes x_r - W_i \otimes x_i - W_j \otimes x_j - W_k \otimes x_k \\
y_i &= W_r \otimes x_r + W_i \otimes x_i + W_j \otimes x_j - W_k \otimes x_k \\
y_j &= W_r \otimes x_r - W_i \otimes x_i + W_j \otimes x_j + W_k \otimes x_k \\
y_k &= W_r \otimes x_r + W_i \otimes x_i - W_j \otimes x_j + W_k \otimes x_k
\end{split}
\end{equation}

This separable formulation represents a valid simplification of the full quaternion convolution. While it relaxes the cross-component interactions (e.g., $W_i \otimes X_j$) present in the complete Hamilton product, it preserves the correct sign patterns essential to quaternion algebra. Our formulation maintains mathematical consistency while achieving the complexity reduction mentioned in Section \autoref{subsection}. Despite the simplification, our experiments demonstrate that this approach effectively captures rotation-aware features while leveraging optimized tensor operations in existing deep learning frameworks. To address computational efficiency, we developed custom CUDA kernels that fuse the Hamilton product operations shown in figure \ref{fig:cuda_kernel}. The backward pass gradients use the same separable structure as the forward pass:

\begin{equation}
\begin{aligned}
\frac{\partial L}{\partial x_q} = \sum_{q \in \{r,i,j,k\}} M_{q}^T \cdot \left(\frac{\partial L}{\partial y_q} \star W_q\right)
\end{aligned}
\end{equation}


where $\mathbf{M}_q^T$ is the mixing matrix derived from the forward pass coefficients in \autoref{eq:5}, \autoref{eq:6}:
\begin{equation}
M = \begin{bmatrix}
1 & 1 & 1 & 1 \\
-1 & 1 & -1 & 1 \\
-1 & 1 & 1 & -1 \\
-1 & -1 & 1 & 1
\end{bmatrix}
\end{equation}

Similarly, for weight gradients:
\begin{equation}
\frac{\partial L}{\partial W_q} = \sum_{q \in \{r,i,j,k\}} M_{q}^T \cdot \left(\frac{\partial L}{\partial y_q} \star x_q\right)
\end{equation}

This fusion reduces memory bandwidth requirements and achieves comparable inference time with standard convolutions.



\subsection{Independent Quaternion Batch Normalization}
Unlike previous quaternion normalization approaches that treat quaternions as unified entities, IQBN normalizes each component independently while preserving important cross-component relationships.

\text{For quaternion feature } $x = \{x_r, x_i, x_j, x_k\}$, $\text{ with } B \times C \times H \times W \times Q \text{ shape}$, where $q \in \{r, i, j, k\}$ represents each quaternion component, IQBN computes:

\begin{equation}
\begin{split}
\mu_q &= \mathbb{E}[X_q] \\
\sigma_q^2 &= \mathbb{E}[(X_q - \mu_q)^2] \\
\hat{x}_q &= \frac{x_q - \mu_q}{\sqrt{\sigma_q^2 + \epsilon}}
\end{split}
\end{equation}




During training, running statistics are computed and maintained:

\begin{equation}
\begin{split}
\mu &= x.\text{mean}([0, 3]) \\
\sigma^2 &= x.\text{var}([0, 3]) \\
x_{\text{norm}} &= \frac{x - \mu.\text{view}(1, C, 1, 1, Q)}{\sqrt{\sigma^2.\text{view}(1, C, 1, 1, Q) + \epsilon}} \\
\text{Output} &= x_{\text{norm}} \cdot \gamma + \beta
\end{split}
\end{equation}

\begin{figure*}[h]
    \centering
    \includegraphics[width=0.75\textwidth]{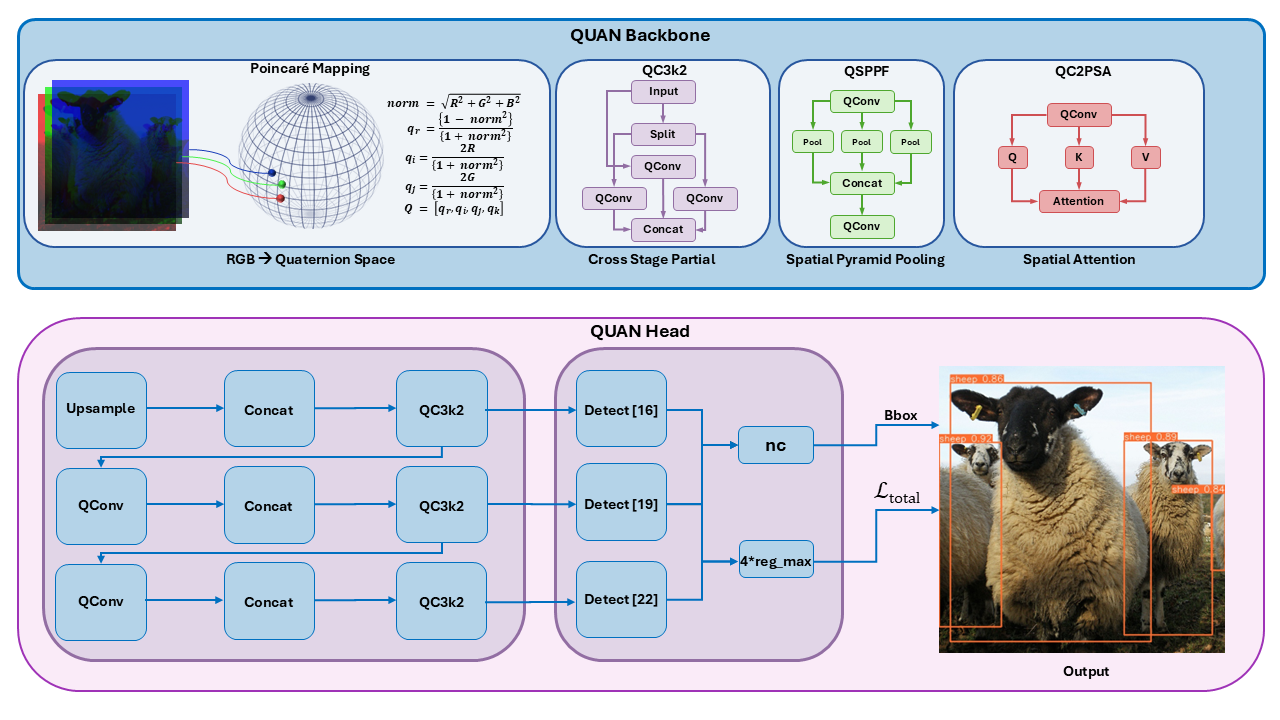}
    \caption{QUAN architecture showing quaternion adaptations of key components: QConv, QC3k2, QSPPF, and QC2PSA}
    \label{fig:qarch}
\end{figure*}

\subsection{Quaternion Architectural Components}
As a novel framework that integrates quaternion approximation with IQBN, QUAN can be leveraged to renovate most CNNs and quaternion CNNs, with necessary modifications to maintain quaternion operations throughout the network structure. In this section, YOLOv11 \cite{yolov11} is renovated with QUAN, and the following modifications, illustrated in Fig. \autoref{fig:qarch}, were made besides replacing standard convolution with approximated quaternion convolution:

\subsubsection{Quaternion Cross Stage Partial with kernel size 2 (QC3k2)}

The C3k2 module balances multi-scale features and gradient flow in YOLO architectures. The quaternion adaptation, QC3k2, maintains the partial cross-stage design while adapting convolutions to quaternion operations:

\begin{equation}
\text{QC3k2}(x) = \text{QConv}\left(\text{Concat}\left(\text{QBottleneck}(x_1), x_2\right)\right)
\end{equation}

where $x_1$ and $x_2$ are the split components of the input feature map.

\subsubsection{Quaternion Spatial Pyramid Pooling (QSPPF)}

SPPF efficiently handles multi-scale features through pyramid pooling. In QSPPF, the quaternion structure is preserved during pooling operations by applying pooling operations independently to each component followed by Hamilton-based mixing. This approach maintains rotation equivariance while capturing multi-scale information.
\subsubsection{Quaternion Partial Spatial Attention (QC2PSA)}

C2PSA enhances the network's ability to focus on relevant spatial regions while maintaining quaternion structure. The module splits input features, applies quaternion attention to one branch, and recombines them. The quaternion attention mechanism computes spatial attention weights while respecting quaternion algebra, ensuring that rotational information is preserved.

\subsection{Loss Functions for Object Detection and Orientation}

For oriented object detection, QUAN employs specialized loss terms that account for both object localization, orientation, and classification:

\begin{equation}
\mathcal{L}_{\text{total}} = \mathcal{L}_{\text{cls}} + \mathcal{L}_{\text{CIoU}} + \lambda_1 \mathcal{L}_{\text{angular}} + \lambda_2 \mathcal{L}_{\text{reg}} + \lambda_3 \mathcal{L}_{\text{smooth}}
\end{equation}

where:
\begin{itemize}
    \item $\mathcal{L}_{\text{cls}} = -\sum_{c=1}^Cy_c \log(\hat{y}_c) + (1 - y_c) \log(1 - \hat{y}_c)$ - the classification loss
    \item $\mathcal{L}_{\text{CIoU}} = 1 - \text{IoU} + \frac{\rho^2(\mathbf{b}, \mathbf{b}^2{gt})}{c^2} + \alpha v$ - the complete IoU loss for bounding box regression
    \item $\mathcal{L}_{\text{angular}} = 2 \arccos (|\langle q_{\text{pred}}, q_{\text{target}}\rangle|)$ - the quaternion angular loss computes the geodesic angular distance between predicted and ground truth quaternion orientations, accounting for the double cover property
    \item $\mathcal{L}_{\text{reg}} = \|q\|_2 - 1^2$ - the quaternion regularization loss enforces unit quaternion constraints to maintain valid rotation representations 
    \item $\mathcal{L}_{\text{smooth}} = \arccos (|\langle q_i, q_{i+1} \rangle|)$ - the orientation smoothness loss encourages consistency between sequential predictions to reduce jitter
\end{itemize}

This integrated loss formulation addresses the challenge of simultaneously optimizing detection accuracy and orientation precision, with the quaternion components designed to preserve the geometric constraints for rotation representation in 3D space.

\section{Experiments}
We evaluate QUAN across two tasks: image classification and object detection. All experiments were conducted on a single NVIDIA RTX 4090 GPU. The experiments are designed to evaluate three key aspects: (1) parameter efficiency compared to standard CNNs and existing quaternion networks, (2) real-world applicability in robotic assembly tasks, and (3) rotation equivariance properties essential for robotic manipulation.

\subsection{Datasets}
\subsubsection{Image Classification Datasets}
We use CIFAR-10/100~\cite{cifar} (60K 32×32 images), and ImageNet~\cite{russakovsky2015imagenetlargescalevisual} subset (100 classes, 130K images).For \textbf{object detection}, we evaluate on COCO 2017~\cite{COCO} (118K training, 5K validation), DOTA-v1.0~\cite{dota} (2,806 aerial images with 188k instances of oriented objects), and a custom robotic workspace dataset\footnote{\url{https://universe.roboflow.com/ukyaism/boxes_2-pwu7p}} (1,900 bids-eye view images with oriented and regular annotations) for pick-and place tasks.


\subsection{Network Architectures}
For image classification, a Quaternion Wide ResNet (QWRN) \cite{wideresnet} was implemented with the 16-4 configuration (16 layers, width factor 4) while YOLOv11n was implemented for detection. These architectures were selected for its efficiency and demonstrated success on state-of-the-art datasets. For classification, we use AdamW optimizer with cosine annealing, batch size 256, initial learning rate (lr) 0.1, 300 epochs. For object detection we primarily use SGD with  batch size 32, automatic lr scheduling, for 100 epochs. Data augmentation includes random crops, horizontal flips, and AutoAugment for classification; mosaic augmentation, and affine transforms for detection.

\subsection{Image Classification Performance}
    We compare QUAN against both standard CNNs and existing quaternion approaches. While several quaternion networks have been proposed~\cite{DQN} \cite{QCNN}, most lack publicly available implementations or are limited to specific tasks. We implemented and evaluated pure quaternion operations (without our approximation), but found latency (3.8× slower) due to the lack of hardware optimization for quaternion arithmetic. For fair comparison, we focus on published results from reproducible quaternion methods and provide our implementation for community evaluation.
Table \ref{tab:cls} presents the classification accuracy across datasets. QUAN achieves state-of-the-art performance among quaternion networks while using significantly fewer parameters.
\begin{table}[h]
    \centering
    \caption{Image classification accuracy and parameter efficiency}
    \begin{adjustbox}{width=\columnwidth}
    \label{tab:cls}
    \begin{tabular}{ccccc}
        \hline
        Model & Params & CIFAR-10 & CIFAR-100 & ImageNet\\
        \hline
        ResNet34 & 3.6M & 94.98\% & 75.97\% & 73.74\%$^*$\\
        WRN-16-4 \cite{wideresnet} & 2.75M & 94.76\% & 76.09\% & 71.09\% \\
        \textbf{QUAN} & \textbf{717K} & \textbf{95.12\%} & \textbf{76.83\%} & \textbf{74.28\%}$^*$ \\
        Deep QCNN \cite{DQN} & 932.8K & 94.56\% & 73.99\% & -\\
        QVGG11 \cite{qactivate} & 3.8M & 85.15\% & - & -\\
        Shallow QCNN \cite{QCNN} & - & 77.78\% & - & -\\
        QResNet34 \cite{Hongo} & 1M & 92-94\% & - & -\\
        \hline
    \end{tabular}
    \end{adjustbox}
    $^*$21.8M parameters for ResNet34, 5.39M parameters for QUAN on ImageNet
\end{table}

QUAN-WRN achieves 95.12\% accuracy on CIFAR-10 with only 717K parameters—74\% fewer than WRN-16-4 while outperforming all quaternion baselines. This parameter efficiency remains at larger scales, where QUAN requires only 5.39M parameters compared to 21.8M for a standard ResNet34, representing an 75\% reduction without sacrificing accuracy. 

This efficiency is valuable for resource-constrained robotic systems, where model size directly impacts deployment feasibility. Despite the dramatic parameter reduction, training time only increased by 6.8\% on average across our classification experiments, confirming that QUAN's approximation approach maintains computational efficiency. 



\subsection{Object Detection Results}

\subsubsection{Classical Object Detection}


Table~\ref{tab:detection_classical} shows QUAN maintains competitive performance on axis-aligned detection while dramatically reducing model size.

\begin{table}[h]
\caption{Classical Object Detection Performance}
\label{tab:detection_classical}
\centering

\resizebox{\columnwidth}{!}{%
\begin{tabular}{llcccc}
\hline
Dataset & Model & Params & mAP50 & mAP50-95 & ms \\
\hline
\multirow{2}{*}{COCO2017} & QUAN-YOLO11n & \textbf{1.187M} & 0.478 & 0.339 & 1.68\\
 & YOLO11n & 2.65M & \textbf{0.547} & \textbf{0.3908} & \textbf{1.5} \\
 \hline
\multirow{2}{*}{Robotic Dataset} & QUAN-YOLO11n & \textbf{0.67M} & 0.989 & \textbf{0.955} & \textbf{1.95} \\
 & YOLO11n & 2.61M & \textbf{0.992} & 0.951 & 2.098 \\
\hline
\end{tabular}%
}
\end{table}

QUAN outperforms standard YOLO11n (0.9548 vs 0.951 mAP50-95) on the robotic dataset while using only 25.3\% of the parameters and 36\% more inference time. On COCO, QUAN underperforms YOLO11n (.339 vs .3908 mAP50-95) with 12\% more inference time. Despite scaling up the architecture, QUAN experiences impaired generalization as the dataset scales likely due to the decreased model size. Additionally, we did not have the computational capacity to use the same batch size and hyperparameters used on the original models when training from scratch. 

\subsubsection{Oriented Object Detection}
For rotation-aware detection critical in robotic applications, Table~\ref{tab:detection_oriented} demonstrates QUAN's effectiveness across aerial and robotic datasets.

\begin{table}[h]
\centering
\caption{Oriented object detection performance}
\label{tab:detection_oriented}
\resizebox{\columnwidth}{!}{%
\begin{tabular}{llcccc}
\toprule
Dataset & Model & Params & mAP@50 & mAP@50-95 & ms \\
\midrule
\multirow{2}{*}{DOTA} & YOLO11n-obb & 2.66M & 77.3 & 61.4 & 4.287\\
& \textbf{QUAN-YOLO11n} & \textbf{1.24M} & 76.2 & 60.8 & 4.61 \\
\midrule
\multirow{2}{*}{Robotic} & YOLO11n-obb & 2.66M & \textbf{93.3} & 68.5 & 3.57 \\
& \textbf{QUAN-YOLO11n} & \textbf{0.69M} & 92.4 & \textbf{76.4} & \textbf{3.44} \\
\bottomrule
\end{tabular}%
}
\vspace{-0.3cm}
\end{table}

\begin{figure}[t]
    \centering
    \includegraphics[width=\columnwidth]{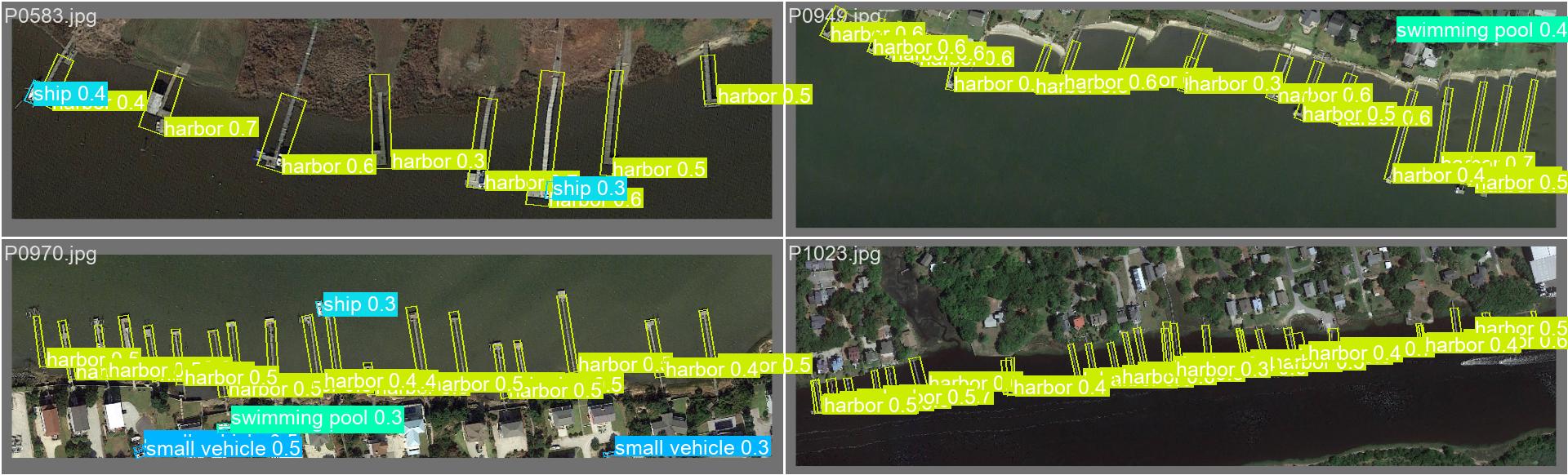}
    \caption{Oriented Object Detection Predictions by QUAN-YOLOv11n}
    \label{fig:obb}
\end{figure}

On the robotic dataset, QUAN outperforms YOLO11n-obb (76.4 vs. 68.5) while reducing the inference time by 3.7\%. Figure \ref{fig:obb} demonstrates QUAN's detection capabilities, showing accurate oriented bounding box annotations. On DOTAv1, we see the same performance impairment (60.8 vs 61.4 map50-95) as we did on COCO likely due to the small model size.





\begin{figure}[t]
    \centering
    \includegraphics[width=.95\columnwidth]{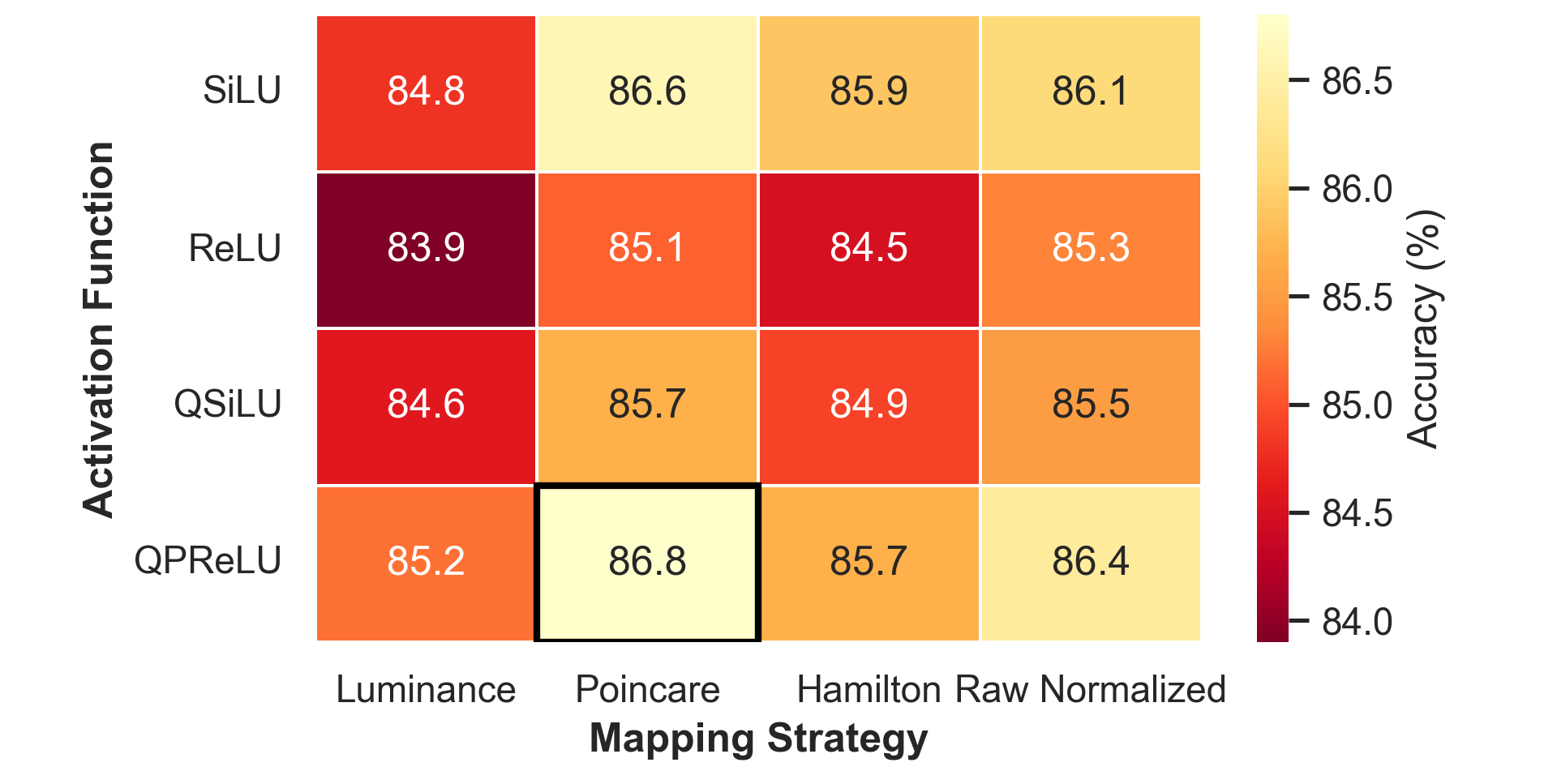}
    \caption{Ablation studies for quaternion activation and mapping}
    \label{fig:heatmap}
\end{figure}

\subsection{Ablation Studies}
Figure \ref{fig:heatmap} presents the ablation study results of quaternion activation and mapping strategies. While Poincaré mapping generally outperforms alternatives across activation functions, SiLU consistently shows superior results over ReLU and QPReLU variants. Traditional quaternion batch normalization showed on average 0.61\% less accuracy than IQBN across the CIFAR-10 experiments and was not included, but may offer advantages in other contexts or with further optimization.

\section{CONCLUSIONS}

Quaternion Approximate Networks (QUAN) advance quaternion neural networks while maintaining practical applicability across both image classification and oriented object detection. By implementing quaternion operations through efficient real-valued approximations and independent normalization, QUAN overcomes key limitations that have hindered the widespread adoption of quaternion networks. The systematic adaptation of standard YOLO architectural blocks (C3k2, SPPF, C2PSA) to quaternion counterparts enables  integration of rotation equivariance into existing object detection frameworks. However, it is not without limitations as the quaternion approximation increases arithmetic operations, resulting in slightly higher inference time that our CUDA kernels partially mitigate. The Poincare mapping, while effective, may not be optimal for all downstream tasks. Future work will explore: self-supervised learning with quaternion representations, optimization over the unique Hamilton matrices to identify optimal task-specific structures, adapting LoRA with quaternions, and live use in robotic assembly applications, where precise spatial understanding is essential for accurate manipulation.

\balance
\bibliographystyle{IEEEtran}
\bibliography{main}
\addtolength{\textheight}{-4cm}   




\vspace{5mm}
\section{APPENDIX}

\subsection{Mapping Strategies}
\label{app:mapping_strategies}

Multiple RGB-to-quaternion mapping strategies have been investigated to determine the optimal approach for preserving both color relationships and geometric properties. The evaluated mappings included:
\begin{itemize}
\item \textbf{Luminance mapping}: Uses perceptual weights (0.299R + 0.587G + 0.114B) for the real component while preserving normalized RGB values in the imaginary components, providing human perception-aligned representations.
    \item \textbf{Mean brightness mapping}: Utilizes the arithmetic mean of RGB channels as the real component, offering computational simplicity.
    \item \textbf{Raw normalized mapping}: Directly normalizes RGB values with equal weighting across channels.
    \item \textbf{Hamilton mapping}: Embeds RGB values as pure imaginary components (0, R, G, B), preserving raw color information.
    \item \textbf{Poincaré mapping}: Projects RGB values onto the unit quaternion sphere, ensuring numerical stability during training while preserving relative color relationships.
\end{itemize}

\end{document}